\documentclass[letterpaper, 10 pt, conference]{ieeeconf}  

\usepackage{graphics} 
\usepackage{graphicx}
\usepackage{amsmath} 
\usepackage{amssymb}  
\usepackage{xcolor}
\usepackage{caption}
\usepackage{multirow}
\usepackage{booktabs}
\usepackage{siunitx}
\usepackage{algorithm}
\usepackage{algorithmic}
\usepackage{hyperref}
\usepackage{cleveref}
\usepackage{caption}
\usepackage{subcaption}
\usepackage{bm}
\usepackage{booktabs}
\usepackage{array}
\usepackage{comment}

\usepackage[T1]{fontenc}
\usepackage{newtxtext,newtxmath}

\usepackage{atbegshi}
\AtBeginShipout{%
  \ifnum\value{page}>1
    \setbox\AtBeginShipoutBox=\vbox{%
      \kern 6pt
      \box\AtBeginShipoutBox
    }%
  \fi
}

\title{\LARGE \bf
Bridging the sim2real gap in the table tennis robot with a transformer-based ball states predictor
}

\author{Yin Bi$^{1}$, Christian Conti$^{2}$, Bilan Yang$^{2}$, Alexander Sigrist$^{1}$, Peter D\"urr$^{1}$, Naoya Takahashi$^{1}$ \\
\\
$^{1}$Sony AI, Z\"urich, Switzerland \\
$^{2}$Sony AI, Tokyo, Japan 
}
\begin{document}

\maketitle

\begin{abstract}

Robotic table tennis is a representative benchmark for high-speed, closed-loop robotic control in dynamic environments, where accurate and fast prediction of ball states is critical for reliable planning and control. Physics-based approaches rely heavily on accurate parameter identification and precise initial state, while learning-based methods often struggle to capture long-range temporal dependencies and are typically trained on limited or simulated data.
We propose a transformer-based framework for table tennis ball state prediction that leverages attention mechanisms to model long-range temporal correlations directly from historical observations, without relying on explicit flight or bounce models. To support robust learning and generalization, we collected a large-scale real-world dataset from players of varying skill levels and diverse ball cannon configurations. The combination of a high-capacity transformer architecture and extensive real-world data enables accurate long-horizon forecasting.
Building on this capability, we introduce a plug-and-play sim-to-real transfer strategy, Swap Predictor at Deployment (SPAD), which replaces the physics-based simulator used during training with the proposed real-world-trained predictor at deployment, improving the sim-to-real transferability of the policy without requiring retraining. We demonstrate that this simple substitution effectively narrows the sim-to-real gap while preserving the efficiency and scalability of simulation-based training.

\end{abstract}

\section{INTRODUCTION}

Table tennis robots have attracted increasing attention since the late 20th century \cite{kyohei2020ping, ding2022goalseye, d2024achieving, hu2025catching, ji2021model, tebbe2021sample, buchler2022learning}, serving as compelling testbeds for integrating perception, mechatronics, control, and learning. 
Successfully returning a fast-moving ball, up to 35 m/s at professional levels \cite{hai2002speed}, requires estimating not only the current ball position but also its future state (position, velocity, and spin). Thus, accurate and fast prediction of ball states is essential for effective planning and control.

\begin{figure}[thpb]
    \centering
    \includegraphics[width=1.0\linewidth]{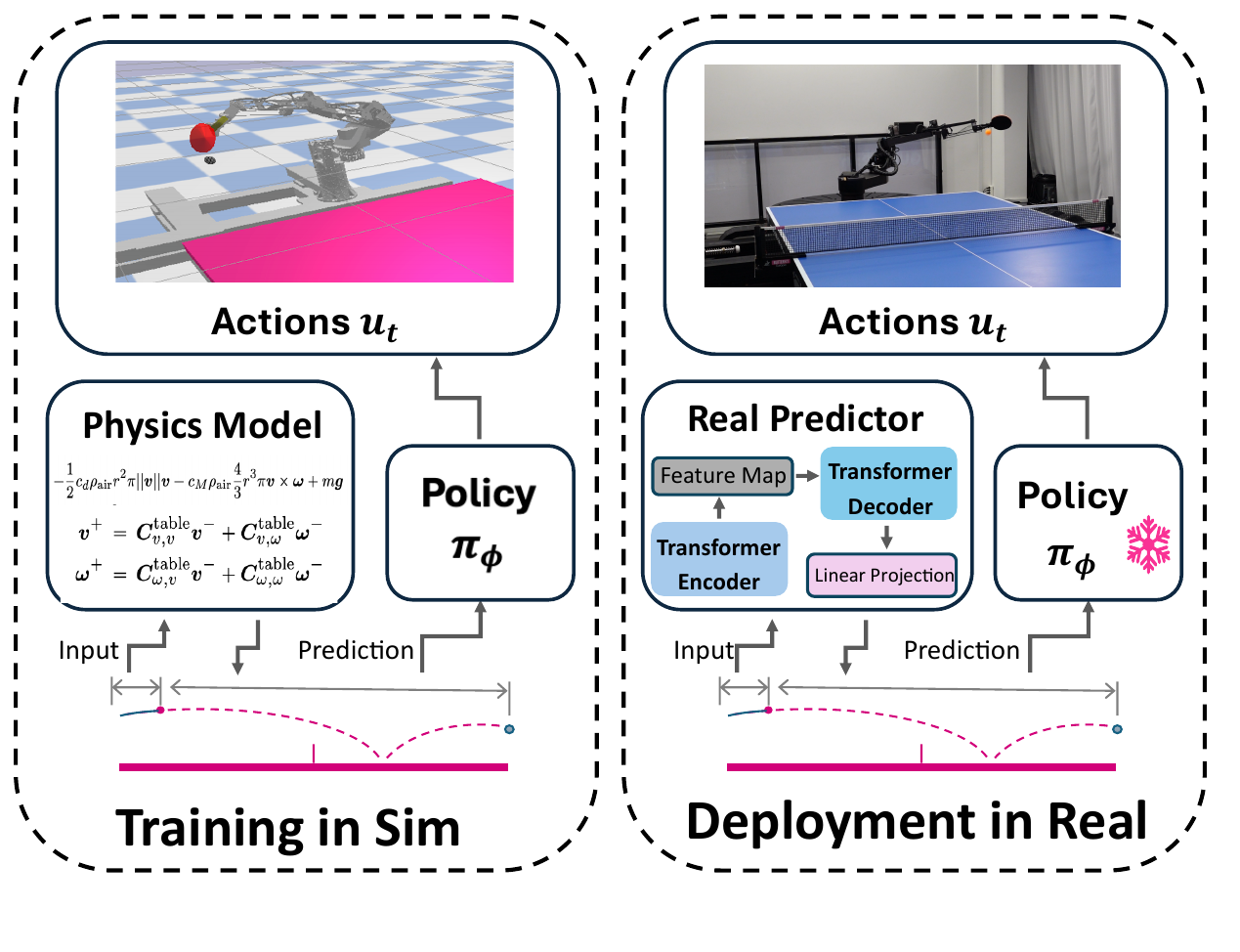} 
    \vspace{-0.7cm}
    \caption{Sim-to-real transfer via predictor replacement: The control policy is first trained using a simulated predictor and then deployed with the real-world-trained predictor, without policy retraining.}
    \label{fig:intro}
    \vspace{-0.3cm}
\end{figure}

Existing physics-based approaches model ball flight and ball-table contact using analytical dynamics with parameters such as drag, restitution, and friction. The current state is estimated from observations using filtering or polynomial fitting \cite{bao2012bouncing, yan2024prediction, su2013trajectory}, and future states are subsequently predicted using physics-based models. While interpretable and physically grounded, these methods rely heavily on accurate parameter identification and initial state estimation. In practice, unmodeled effects, such as time-varying spin, complex non-linear properties of the table and ball, air disturbances, and contact uncertainties, can degrade accuracy.

In contrast, learning-based approaches learn ball flight and ball-table contact dynamics directly from data, typically using neural networks \cite{he2024modeling, toussaint2025real, li2025spatiotemporal}. They offer greater flexibility and aim to capture complex aerodynamic and contact dynamics implicitly. However, these methods often struggle to model long-range temporal dependencies and are trained on limited real-world datasets or only on simulated data, which can lead to overfitting and sim-to-real discrepancies. Most existing works focus on position prediction, leaving spin - critical for trajectory curvature and rebound - largely underexplored. The absence of standardized datasets and evaluation protocols further complicates fair comparison across methods.

To address these challenges, we propose a transformer-based predictor for predicting table tennis ball states. Inspired by the success of attention mechanisms in modeling long-range dependencies \cite{vaswani2017attention}, our predictor similarly leverages the attention mechanisms to capture temporal correlations over extended observation horizons without relying on analytical flight or bounce models. Specifically, the predictor uses an encoder–decoder architecture with multi-head attention and expansion-bottleneck convolutional blocks, taking historical ball positions and spins as input and forecasting future states (positions, velocities, and spins).

To ensure robustness and real-world generalization, we collected a large-scale training dataset of over 100k ball trajectories, including players of varying skill levels and ball cannons with diverse configurations. Experimental results show that our predictor outperforms physics-based and learning-based baselines, particularly in long-horizon prediction. We will release the testing dataset upon acceptance to enable standardized comparisons in table tennis ball state prediction.

Training control policies in simulation is common in table tennis robotics due to its safety, scalability, and efficiency \cite{buchler2022learning, d2024achieving, ma2023reinforcement}. In typical simulation pipelines, ball trajectories are generated using physics-based models, from which future ball states can be directly derived as policy inputs, making training convenient and computationally efficient. However, inaccuracies in ball dynamics, bounce behavior, and sensor models create a sim-to-real gap that limits real-world performance. Existing strategies, such as system identification \cite{zhu2017model, kaspar2020sim2real, sobanbabu2025sampling}, domain randomization and adaptation \cite{tobin2017domain, loquercio2019deep, ganin2016domain, chebotar2019closing}, and incorporating realistic sensor and actuator noise models into simulation \cite{peng2018sim, tan2018sim, ota2020towards}, can mitigate this gap but are often complex, computationally expensive, or require retraining that is impractical when real-world data are limited.

We propose a simple, plug-and-play sim-to-real transfer strategy, Swap Predictor at Deployment (SPAD): after training the control policy entirely in simulation, the physics-based simulator is replaced with a real-world-trained predictor at deployment, as illustrated in Fig.~\ref{fig:intro}. This substitution requires no policy retraining. Experimental results demonstrate that the proposed approach significantly mitigates the sim-to-real gap. As a result, it preserves the efficiency of simulation-based training while enhancing real-world performance. This strategy may also be applicable to other robotic systems in which predictive models provide inputs to downstream control policies

The main contributions of this work are as follows:
\begin{itemize}
\item We propose a transformer-based framework for predicting table tennis ball states - including ball positions, velocities, and spins - that outperforms physics-based and learning-based baselines, particularly for long-horizon predictions.
\item We collect a large-scale real-world dataset containing 100k ball trajectories with spin measurements, and release a benchmark dataset for standardizing the evaluation protocol.
\item We propose and validate a simple yet effective sim-to-real transfer strategy, SPAD, which enables efficient simulation-based training while mitigating the sim-to-real gap. 
\end{itemize}

\section{Related Work}

\subsubsection{\textbf{Table tennis ball state prediction}}

Physics-based methods model the ball aerodynamics in flight and ball-table contact using analytical equations with estimated physical parameters\cite{nakashima2009modeling, bao2012bouncing, zhao2016rebound, nakashima2010modeling}. Future ball states are predicted using these models, with the initial states estimated from observations via polynomial fitting or filtering techniques such as the Extended Kalman Filter (EKF) or the Unscented Kalman Filter (UKF)\cite{bao2012bouncing, yan2024prediction, su2013trajectory}. While physically interpretable, these methods rely on accurate parameter and initial state estimation, and unmodeled effects, such as air disturbances and surface uncertainties, can degrade performance in real-world deployment. 

Learning-based approaches aim to model ball flight and ball-table contact dynamics directly from data, without explicit physics knowledge. Neural networks are often used to predict trajectories before and after bounces, to acquire the hitting point or landing plane \cite{lin2020ball, lin2019ball, liu2022application}. And recurrent architectures are also widely used for efficiently capturing temporal dependencies \cite{he2024modeling, lin2019simulation, toussaint2025real}. Sequence-to-sequence and trajectory autoencoder models \cite{gomez2020real, toussaint2025real, gao2022model} further mitigate error accumulation in long-horizon prediction by preventing predicted outputs from being recursively fed back into the network as inputs.
Despite these advances, most methods predict only position, neglecting velocity and spin, and rely on small or simulated datasets, limiting real-world generalization. For example, \cite{lin2020ball} uses only 200 real-world and 330 simulated trajectories, which is insufficient to capture the full range of ball dynamics, including velocities up to 35 m/s and angular velocities up to 1000 rad/s \cite{hai2002speed}. Training on synthetic data generated from physics models also inherits the same limitations when deployed on real robots.

Hybrid methods combine physics-based models with learning components, using neural networks to estimate parameters or model deviations while preserving physical plausibility. For example, \cite{achterhold2023black} uses a gray-box model to learn both system dynamics and EKF parameters, while \cite{toussaint5025167hybrid} integrates a differentiable physics model with GRU-based trajectory encoders. Although these approaches improve prediction accuracy, they are often trained on small datasets (e.g., 2000 simulated and 300 real trajectories in \cite{toussaint5025167hybrid}) and still face generalization challenges.

\subsubsection{\textbf{Sim-to-Real Transfer}}

Sim-to-real transfer seeks to enable control policies trained in simulation to perform reliably on physical robots. Common strategies include system identification, domain randomization, and domain adaptation.
System identification attempts to reduce the sim-to-real gap by accurately modeling the physical properties of robots and their environments in simulation \cite{zhu2017model, allevato2020tunenet, kaspar2020sim2real, sobanbabu2025sampling}. While effective, it is sensitive to unmodeled effects and may not generalize well across tasks or environments.
Domain randomization improves robustness by varying simulation parameters such as object appearance, textures, lighting, physics properties, and sensor noise \cite{tobin2017domain, peng2018sim, loquercio2019deep}. Domain adaptation explicitly aligns simulated and real-world data distributions via feature alignment, adversarial training, or fine-tuning on limited real data \cite{ganin2016domain, chebotar2019closing}.
Differences in sensing, actuation, and computational latency between simulation and real robots also affect transfer performance. 
Incorporating realistic noise, latency, and actuation models can reduce the gap \cite{tobin2017domain, peng2018sim, tan2018sim, ota2020towards}, although coverage of real-world variability remains limited.

In robotic table tennis, \cite{buchler2022learning} proposes a hybrid procedure that relies on real-world data whenever possible while using simulation to supplement scarce samples. \cite{d2024achieving} initializes simulation parameters via system identification, applies domain randomization, and uses curriculum learning to refine policies with real-world data. The iterative approach in \cite{abeyruwan2023sim2real} alternates between simulation training and real-world deployment, using human-robot interaction data to refine simulators, but requires repeated real-world data collection. Zero-shot strategies \cite{d2023robotic, Durr2026ace} rely on high-fidelity simulation to deploy policies without real-world fine-tuning, although they remain sensitive to modeling errors under distribution shifts.

\section{Methods}
\subsection{table tennis ball state predictor}

\subsubsection{Architecture of Predictor}

\begin{figure}[thpb]
    \centering
    \includegraphics[width=0.85\linewidth]{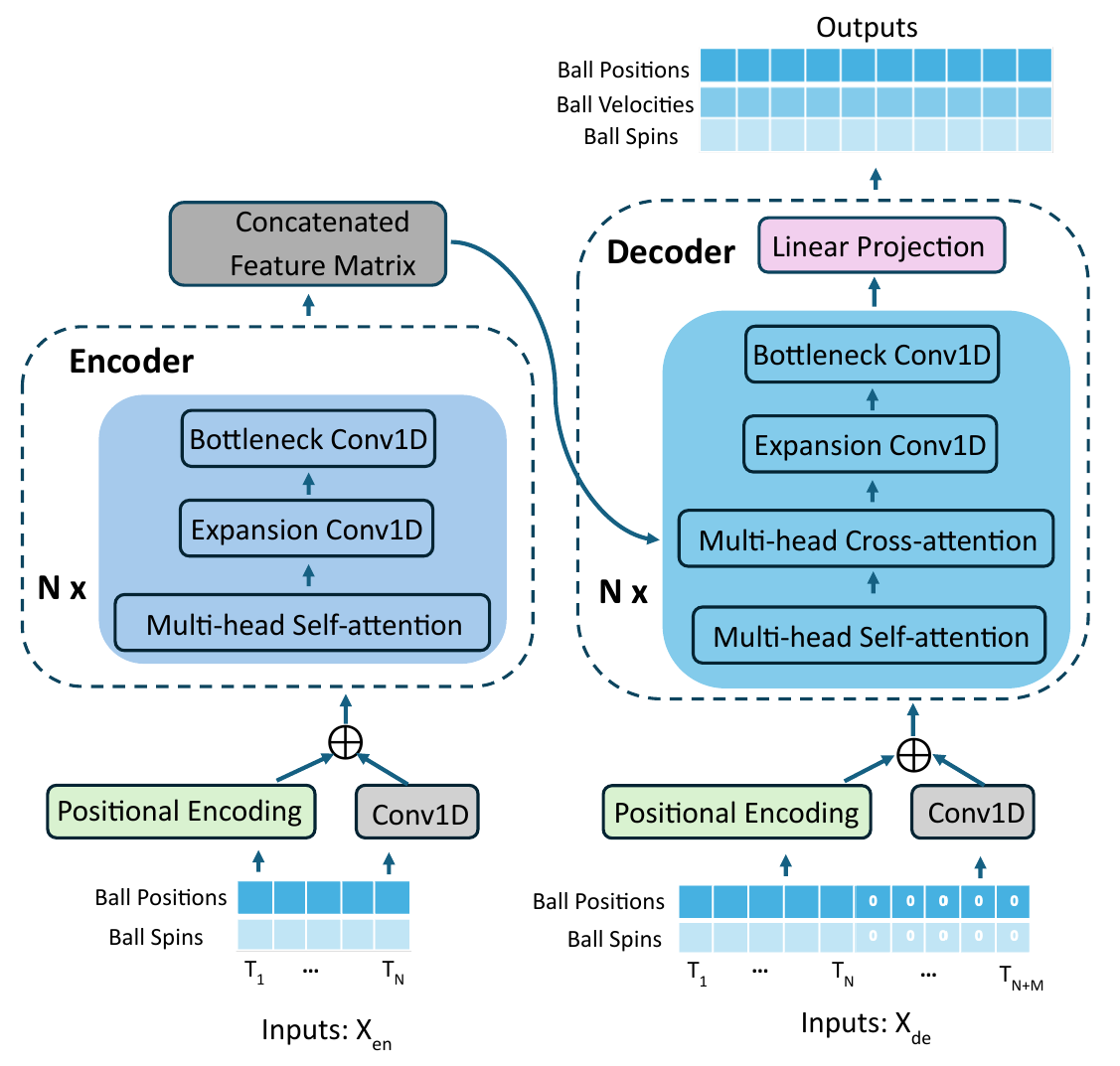} 
    \caption{Overview of the table tennis ball states predictor.}
    \label{fig:archi}
    \vspace{-0.3cm}
\end{figure}

Following the approaches in \cite{zhou2021informer, wu2021autoformer}, we adopt an encoder-decoder architecture for our predictor. This architecture generates long temporal sequences in a single forward pass instead of iteratively, significantly improving inference speed while reducing error accumulation from autoregressive conditioning, thereby preserving accuracy over long horizons. 
An overview of the adopted architecture is shown in Fig.~\ref{fig:archi}. 

The encoder captures temporal correlations in the input sequence and encodes them into a structured feature matrix. This representation is subsequently utilized by the decoder to refine the predicted future ball states by incorporating information from previously passed balls. Following the design of \cite{vaswani2017attention}, the encoder consists of $L_{en}$ identical layers, each comprising three sub-layers. The first sub-layer is a multi-head self-attention mechanism, which is followed by a convolution block including two consecutive 1D convolution layers. The first 1D convolution layer expands the feature dimension to a higher channel space $d_\text{ff}$, and the second projects it back to a lower-dimensional representation, i.e., $d_\text{model}$. The attention mechanism is the standard Scaled Dot-Product Attention \cite{vaswani2017attention}. Residual connections and layer normalization are applied to the self-attention layer and the convolution block. The computations of the $l$-th encoder layer are summarized as follows:
\begin{equation}
\begin{split}
\mathbf{F_{en}^{l,att}} = \text{LayerNorm}\big(\text{Self\_Attention}(\mathbf{X_{en}^{l}} ) + \mathbf{X_{en}^{l}} \big) \\[2mm]
\mathbf{F_{en}^{l}} = \text{LayerNorm}\big(\text{Conv1D}(\text{Conv1D}(\mathbf{F_{en}^{l,att}} )) + \mathbf{F_{en}^{l,att}} \big)
\end{split}
\end{equation}
where $\mathbf{X_{en}^{l}}$ and $\mathbf{F_{en}^{l}}$ denote the input and output of the $l$-th encoder layer, respectively. $\mathbf{F_{en}^{l,att}}$,  represents the output of the self-attention sub-layer in the $l$-th encoder layer.

Similarly, the decoder consists of $L_{de}$ identical layers followed by a final linear projection layer that maps the model representation from dimension $d_\text{model}$ to 9. Compared with the encoder, each decoder layer inserts an additional multi-head cross-attention sub-layer between the self-attention sub-layer and the convolution block. This cross-attention mechanism attends to the encoder outputs using the decoder states as queries. All sub-layers in the decoder are equipped with residual connections and layer normalization, following the same design principles as the encoder.

\subsubsection{Input Representation} 
The model takes a history of $N$ ball positions and spins as inputs and predicts $M$ future ball positions, velocities, and spins.
At time $T$, $N$ historical ball positions and spins are concatenated to form the encoder input:
\begin{equation}
\mathbf{X}_{\text{en}} = (\mathbf{p}, \mathbf{\omega}) \in \mathbb{R}^{N \times 6},
\end{equation}
where $\mathbf{p}\in \mathbb{R}^{N \times 3}$ and $\mathbf{\omega} \in \mathbb{R}^{N \times 3}$ correspond to the $N$ historical 3D position and spin vectors of the ball, respectively.  

For the decoder input, the sequence length is extended to $N+M$ by concatenating $M \times 6$ zero vectors to $\mathbf{X}_{\text{en}}$. These zero vectors serve as placeholders for the future $M$ step, enabling one-step-forward inference rather than iterative prediction. For both the encoder and decoder inputs, if the observed position or spin is unavailable in practice, the corresponding entries are set to zero.

To preserve the local temporal context of the observations, we employ a standard sinusoidal positional encoding \cite{vaswani2017attention} to distinguish the order of observations. This produces a positional encoding matrix 
$\mathbf{P} \in \mathbb{R}^{\text{N} \times d_\text{model}}$ added to the latent feature representation to inject temporal information.

To align the feature dimensions, the observations are projected into a latent feature space using a one-dimensional convolutional layer (with a kernel size of 3 and a stride of 1) after the normalization.
\begin{equation}
\mathbf{F} = \text{Conv1D}(\text{Norm}(\mathbf{X_{en}}), \, k=3, \, s=1)
\end{equation}
The final input representation $\mathbf{R} $ is then obtained by combining the projected features with the positional embeddings: $\mathbf{R} = \mathbf{F} + \mathbf{P}$.

\subsection{Sim2Real Transfer}
A common approach to training control policies for robotic table tennis is to rely on simulated environments for policy learning \cite{abeyruwan2023sim2real, d2024achieving, ma2023reinforcement, buchler2022learning}, as simulation enables efficient parallelization and scalable data generation. However, policies trained exclusively in simulation often fail to generalize to real-world settings due to the domain gap between synthetic and real data. To mitigate this issue, we incorporate a predictor trained on real-world data to provide more accurate estimates of future ball states as inputs to the policy, thereby reducing the sim-to-real discrepancy. 

At each time step $t$, the robot observes the recent historical ball states $o_{1:t}$ and its own state $s_t$. 
A real ball-state predictor $\mathcal{P}_{\text{real}}$ estimates the future ball states over a prediction horizon $[t+1, t+H]$:
\begin{equation}
\hat{X}_{t+1:t+H} = \mathcal{P}_{\text{real}}(o_{1:t}).
\end{equation}
The control policy $\pi_\phi$ then generates the action conditioned on both the robot state and the predicted future ball states:
\begin{equation}
u_t = \pi_\phi(s_t, \hat{X}_{t+1:t+H}).
\end{equation}

During training, future ball states are generated by a physics-based simulator $\mathcal{P}_{\text{sim}}$, due to its efficiency and scalability, which incorporates analytical models of ball flight and ball-table contact dynamics. The simulator is initialized from diverse ball states to cover a wide range of task configurations. The policy is trained entirely in simulation to maximize task reward (e.g., successfully returning the ball) using the robot state and the simulator-generated future ball states as inputs.

The primary sim-to-real discrepancy arises from mismatches between the physics-based simulator $\mathcal{P}_{\text{sim}}$ used during training and the actual physical dynamics of the real world, where in practice $\mathcal{P}_{\text{sim}}$ is conditioned on the initial state estimation and also does not perfectly match reality due to unmodeled aerodynamics, spin-dependent effects, contact dynamics, and environmental disturbances. 
As a result, $\mathcal{P}_{\text{sim}}$ produces incorrect future ball states when applied to real-world observations. Since the policy conditions on future ball states rather than raw sensory inputs, the sim-to-real gap manifests as a distribution mismatch in policy input space:
\begin{equation}
\mathcal{P}_{\text{sim}}(o_{1:t}) \neq \mathcal{P}_{\text{real}}(o_{1:t}).
\end{equation}
Consequently, the action $u_t$ generated by the policy may deviate from the optimal real-world control behavior.

To mitigate this discrepancy, we correct the policy input, i.e., the future ball states. After training, we obtain a fixed policy $\pi_{\phi^*}$. During deployment, we replace the physics-based simulator $\mathcal{P}_{\text{sim}}$ with the real-world-trained predictor $\hat{\mathcal{P}}_{\text{real}}$, while keeping the control policy $\pi_{\phi^*}$ unchanged. We refer to this approach as Swap Predictor at Deployment (SPAD).
The real-world-trained predictor is expected to represent the physical dynamics of the real world more faithfully; this design aligns the distribution of predicted ball states fed into the policy. Instead of adapting the downstream control strategy, we correct the upstream prediction bias introduced by simulation inaccuracies. Once sufficiently accurate future ball states are supplied, the high-level control policy learned in simulation remains effective in the real world. Therefore, predictor replacement alone is sufficient to preserve policy performance without policy retraining. 

Algorithm~\ref{alg:sim2real} summarizes the SPAD procedure for sim-to-real transfer. The approach is modular and broadly applicable to robotic systems in which predictive models are used to provide structured inputs to downstream control policies.

\begin{algorithm}[t]
\caption{Sim2Real Transfer via SPAD}
\label{alg:sim2real}
\begin{algorithmic}[1]
\STATE \textbf{Phase 1: Simulation Training}
\STATE Train control policy $\pi_{\phi^*}$ in simulation using predictions from physics-based simulator $\mathcal{P}_{\text{sim}}$
\vspace{0.5em}
\STATE \textbf{Phase 2: Real-World Adaptation}
\STATE Collect real-world ball dataset $\mathcal{D}_{\text{real}}$
\STATE Train real-world predictor $\hat{\mathcal{P}}_{\text{real}}$ using supervised learning on $\mathcal{D}_{\text{real}}$
\vspace{0.5em}
\STATE \textbf{Deployment (No Policy Retraining)}
\STATE Replace predictor: $\mathcal{P}_{\text{sim}} \rightarrow \hat{\mathcal{P}}_{\text{real}}$
\FOR{each real-world rally}
    \STATE Observe ball state $o_{1:t}$ and robot state $s_t$
    \STATE Predict future ball states $\hat{X}_{t+1:t+H} = \hat{\mathcal{P}}_{\text{real}}(o_{1:t})$
    \STATE Execute action $u_t = \pi_{\phi^*}(s_t, \hat{X}_{t+1:t+H})$
\ENDFOR
\end{algorithmic}
\end{algorithm}

\section{Evaluation and Results}


\subsection{Implementation and Evaluation of Predictor}
\subsubsection{Dataset} \label{sec:dataset}
The dataset used to train the ball state predictor is collected from real-world settings and consists of two categories: game data and ball-cannon data.
For the game data, amateur, club-level, semi-professional, and professional players are invited to participate in data collection sessions. Players are paired and instructed to play competitive matches, aiming to win against their opponents. 
For the ball cannon data, a ball cannon launches balls from predefined positions with varied yaw and pitch angles, initial velocities, and spin rates. Various combinations of these parameters are systematically sampled to increase the diversity of the ball-cannon dataset. 
During data collection, ball positions and spins are recorded. Six Active Pixel Sensor (APS) cameras are installed around the court to ensure full coverage of the playing area. These cameras locate the ball’s 3D position using a color-based filtering detection algorithm \cite{nummiaro2003adaptive}, with an error of 3 mm. Predefined dot patterns are applied to the surface of the ball, and the ball’s orientation is reconstructed from these patterns using geometric hashing \cite{wolfson2002geometric}. The average error of the spin is approximately 5\% of its amplitude. Both ball positions and spins are recorded at a sampling frequency of 200 Hz. Ball velocity is then estimated from position data. 
All recordings are segmented into episodes: for the game data, each episode starts from a racket–ball contact and ends either at rally termination or at the next racket–ball contact; for the ball cannon data, each episode corresponds to a single ball shot. In total, approximately 50k game episodes and 55k ball cannon episodes are collected. Among them, 500 episodes comprising both game and ball cannon episodes are reserved for testing, and the remaining episodes are split into training and evaluation sets with a 9:1 ratio.
The testing episodes will be made publicly available upon acceptance to standardize the dataset used in the evaluation of the table tennis ball state predictor. A small subset of these episodes is provided in the \textbf{supplementary material} for review.

\subsubsection{Implementation Details} \label{sec:implementation}
The predictor forecasts 60 future ball states (M=60) based on 12 historical ball states (N=12). 
The encoder comprises three identical layers ($L_{en}=3$), while the decoder consists of two identical layers ($L_{de}=2$). The feature dimension of the attention layers is set to 512 ($d_{model}=512$), and the expanded dimension of the convolution layers in both the encoder and decoder is 1024 ($d_{ff}=1024$). All attention layers employ 8 attention heads. During training, the batch size is set to 128, and the predictor is trained for 80 epochs using the Adam optimizer \cite{kingma2014adam}. The loss is computed as the mean squared error (MSE) loss over the prediction horizon between the predicted and ground-truth ball states. The initial learning rate is 0.0001 and decays by a factor of 0.95 per epoch.

\subsubsection{Evaluation Results} 
\begin{figure}[thpb]
    \centering
    \includegraphics[width=0.96\linewidth]{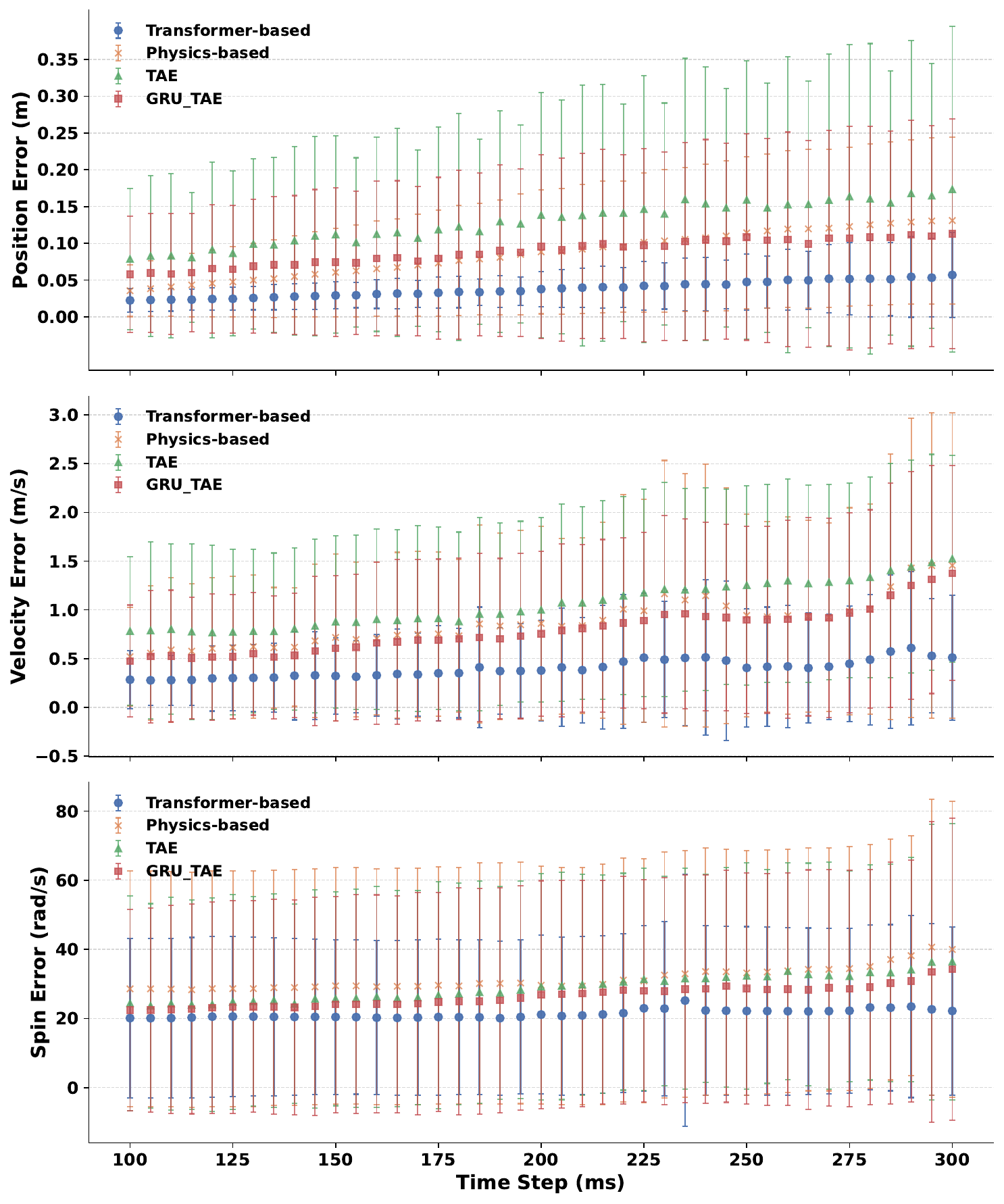} 
    \caption{Prediction accuracy of ball state (position, velocity, and spin) across horizons from 100-300 ms, reporting mean error and standard deviation for the proposed predictor and the physics-based, TAE, and GRU-TAE baselines.}
    \label{fig:accuracy_predictor}
    \vspace{-0.3cm}
\end{figure}

We compare the proposed predictor against a physics-based method and learning-based approaches, including the Trajectory Autoencoder (TAE) and the gated recurrent unit TAE (GRU-TAE). For the physics-based method, an Extended Kalman Filter (EKF) \cite{simon2006optimal} is first used to estimate the initial state based on the first 12 historical ball states. Subsequently, future states are predicted by propagating the estimated state through ball flight \cite{nakashima2010modeling} and ball-table contact \cite{nakashima2011robotic} models. 
For TAE and GRU-TAE, we extend the implementations in \cite{toussaint2025real} by increasing the input and output dimension from 3 (position-only) to 6 and 9, respectively. This modification enables the models to incorporate both ball position and spin as inputs and to simultaneously predict ball position, velocity, and spin. Both are trained using the same training procedure and the same training and evaluation datasets described in \Cref{sec:implementation} and \Cref{sec:dataset}.
Prediction accuracy is evaluated over horizons from 100 ms to 300 ms with 5 ms temporal resolution. 
The prediction error is defined as the Euclidean norm of the difference between the predicted state and the corresponding ground-truth state at each prediction time step. Fig.~\ref{fig:accuracy_predictor} summarizes the mean error and standard deviation across all testing episodes.
As shown in Fig.~\ref{fig:accuracy_predictor}, the proposed transformer-based ball state predictor exhibits consistently lower mean prediction error throughout the evaluated horizon. 
In addition, the error variance is small over the prediction horizon, indicating stable long-term prediction behavior. 
\subsubsection{Real-world Deployment} \label{sec:real_world_deploy}
To evaluate real-world performance, the proposed predictor was integrated into a physical robotic platform for human–robot table tennis rallies. Raw ball positions and spin measurements from external sensors are provided to the predictor, which generates future ball states from 5–300 ms at 5 ms intervals. These predictions are used as inputs to a control policy governing the robotic arm. The policy is a deep reinforcement learning (RL) policy trained entirely on single shots in simulation using the Soft Actor-Critic (SAC) algorithm \cite{haarnoja2018soft}, which learns to defeat the opponent with a set of desired skills using Universal Value Function Approximators \cite{schaul2015universal}.
A demonstration video showing the integration of our ball-state predictor with a real system for human-robot rallies is included in the \textbf{supplementary material}.

\subsection{Sim-to-real Transfer}
We evaluate the proposed SPAD sim-to-real transfer method under two settings: sim-to-sim and sim-to-real. In the sim-to-sim setting, we assess policy transfer between two simulated domains with differing physical parameters, enabling a controlled analysis of domain discrepancies. In the sim-to-real setting, a policy trained in simulation is evaluated on real-world data, the primary interest of this work.

\subsubsection{Proof of Concept - Sim-to-sim}  \label{sec:sim2sim}
The sim-to-sim experiments serve as a proof of concept in which the policy is trained in one simulation environment and evaluated in another with modified physical parameters. As described in \Cref{sec:real_world_deploy}, the control policy is trained with deep reinforcement learning using the Soft Actor-Critic algorithm \cite{haarnoja2018soft} to defeat the opponent with different desired skills via Universal Value Function Approximators \cite{schaul2015universal}. For the ball simulator, established physics-based models are used to describe both ball flight \cite{nakashima2010modeling} and ball–table contact \cite{nakashima2011robotic}. 
Parameters of ball physics models are varied to construct different simulators. Among the key parameters requiring identification are the aerodynamic drag coefficient for ball flight and the table restitution coefficient for ball-table contact. 
We therefore set a drag coefficient to 0.55 and a table restitution coefficient to 0.98 to construct a pseudo "ground-truth" simulator. To emulate modeling inaccuracies, we construct "imperfect" simulators by varying these two parameters. Specifically, three types of imperfect ball simulators are considered:
(1) "Imperfect A": only the table restitution coefficient in the ball-table contact model is modified to 0.9;
(2) "Imperfect B": only the drag coefficient for ball flight is modified to 0.4;
(3) "Imperfect C": both the drag coefficient and the table restitution coefficient are modified to 0.4 and 0.9, respectively. 

\begin{figure}[thpb]
    \centering
    \includegraphics[width=0.98\linewidth]{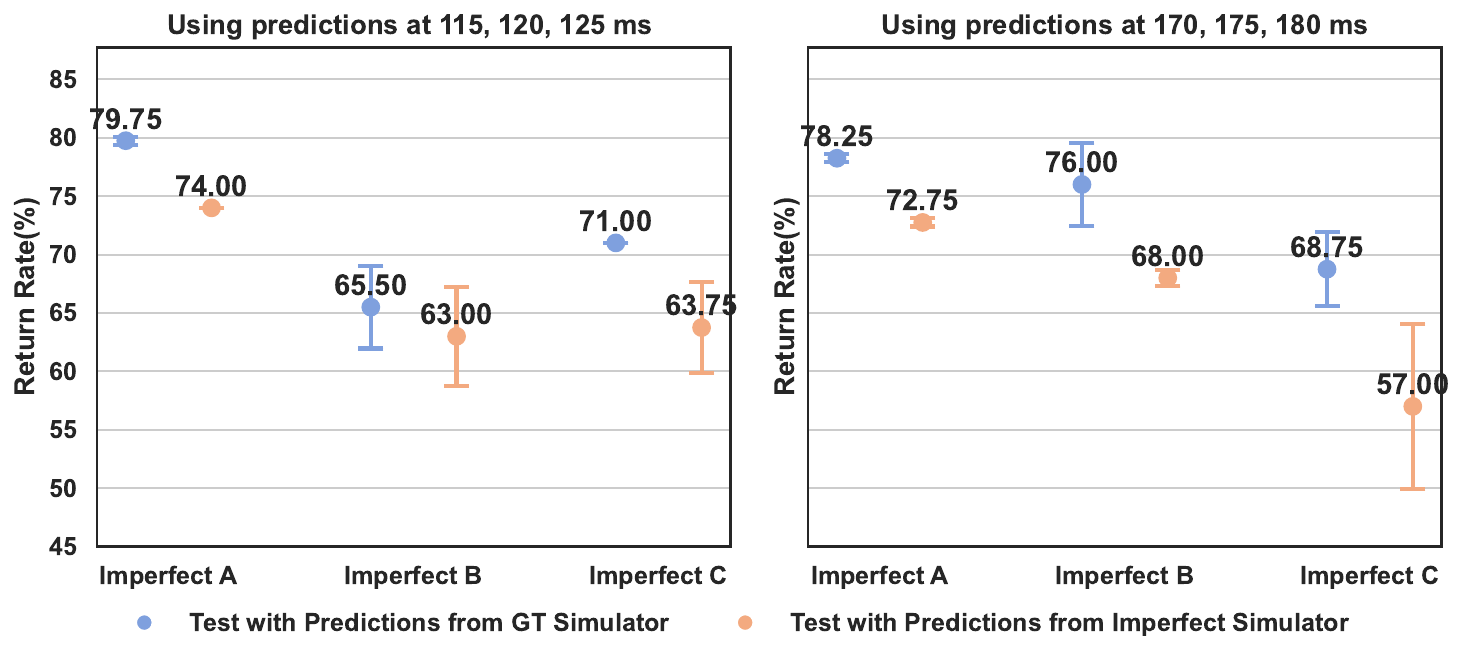} 
    \caption{Sim-to-sim transfer: Policies are trained using ball trajectories and future-state predictions generated by three "imperfect" simulators (A-C), respectively. At test time, trajectories are generated by the "ground-truth" simulator, while future states provided to the policy are predicted by either the "ground-truth" simulator (blue) or the "imperfect" simulator used during training (orange). Return rates are reported for two prediction horizons (115–125 ms and 170–180 ms). Improved performance with ground-truth predictions indicates effective transfer.}
    \label{fig:sim2sim}
    \vspace{-0.2cm}
\end{figure}

\begin{table*}[thpb]
\centering
\caption{Sim-to-real transfer results with respect to Stroke-Uni, Stroke-Pro, and Receive.}
\label{tab:sim-to-real_results}
\setlength{\tabcolsep}{8pt} 

\begin{tabular}{>{\centering\arraybackslash}m{3cm} cccccccc}
\toprule
\multirow{2}{*}{} & \multicolumn{4}{c}{Using predictions at 115, 120, 125 ms} & \multicolumn{4}{c}{Using predictions at 170, 175, 180 ms} \\
\cmidrule(lr){2-5} \cmidrule(lr){6-9}
 & \multicolumn{2}{c}{Physics-based Predictor} & \multicolumn{2}{c}{Real-world Predictor} & \multicolumn{2}{c}{Physics-based Predictor} & \multicolumn{2}{c}{Real-world Predictor} \\
\cmidrule(lr){2-3} \cmidrule(lr){4-5} \cmidrule(lr){6-7} \cmidrule(lr){8-9}
 & Hits & Returns & Hits & Returns & Hits & Returns & Hits & Returns \\
\midrule
Stroke-Pro & 136 & 81 & 170 & 117 & 114 & 75 & 164 & 92 \\
Stroke-Uni & 154 & 80 & 195 & 146 & 137 & 86 & 187 & 125 \\
Receive & 98 & 68 & 98 & 76 & 96 & 67 & 97 & 68 \\
\bottomrule
\end{tabular}
\end{table*}

During policy training, the imperfect ball simulator is used to generate ball trajectories from sampled initial ball states. These trajectories represent the balls that the policy is required to return. A short-term horizon of future ball states from the generated trajectories is provided as part of the input to the policy. 
To evaluate sim-to-sim transfer, test trajectories are generated using the "ground-truth" simulator, and the policy is tasked with returning these balls. Policy performance is evaluated under two conditions: given the first ball state from each test trajectory, future ball states provided to the policy are predicted using either (1) the "ground-truth" simulator or (2) the imperfect simulator (as used during training). Comparing performance under these two conditions allows us to assess transfer between the two simulated domains.

Performance is measured by the return rate, defined as the percentage of balls successfully returned to the opponent’s side of the table. Each evaluation consists of 300 episodes. Results are shown in Fig.~\ref{fig:sim2sim}. To ensure robustness, we evaluate two prediction horizons by training policies using predicted future ball states at [115, 120, 125] ms and [170, 175, 180] ms as inputs, respectively, consistent with reported planning and reaction latencies of 100-200 ms in dynamic robotic systems \cite{zabalza2019smart}. For each setting, the policy is trained twice to account for stochasticity in training and the mean and variance across these two runs are reported. The results consistently show that when the incoming ball is generated by the "ground-truth" simulator, policies that use future ball states predicted by the "ground-truth" simulator as inputs achieve a higher return rate than those using predictions from the imperfect simulator, even though the policies were trained with an imperfect simulator to generate both the training ball trajectories and the corresponding future ball state predictions. This trend is observed across all three imperfect simulator settings, supporting the effectiveness of predictor replacement for mitigating domain discrepancies.

\subsubsection{Real-World Data Testing - Sim-to-real}

In the sim-to-real evaluation, the control policy is trained entirely in simulation and then evaluated using real-world ball data, allowing us to assess how effectively the proposed framework bridges the sim-to-real gap. 
For real-world data evaluation, we collect a diverse test set consisting of 200 rally shots from university-level players (Stroke-Uni), 200 rally shots from professional-level players (Stroke-Pro), and 100 serves (Receive) from both groups. 

During training, the same control policy and ball simulator described in \Cref{sec:sim2sim} are used. In the ball simulator, the aerodynamic drag coefficient and the table restitution coefficient are estimated as variable parameters \cite{miyazaki2017lift, ittf_statutes}. These parameters are fixed at representative values within their ranges: the drag coefficient is set to 0.4 and the table restitution coefficient to 0.9.
During evaluation, each real-world trajectory is replayed to the policy, which uses future ball states as its input. The future ball states are predicted either by the physics-based predictor or by the real-world predictor. To quantify sim-to-real transfer performance, we compare the number of successful hits and returns achieved under these two predictor settings. Table \ref{tab:sim-to-real_results} reports the results for using two different prediction horizons as ball state input (predictions at [115, 120, 125] ms and at [170, 175, 180] ms). Across both prediction horizons and all datasets, using a real-world predictor consistently improves performance over the simulator in terms of the number of hits and returns, especially in the Stroke-Uni shots, illustrating that the proposed ball states predictor can alleviate the sim-to-real gap. 

To further evaluate the quality of the shots returned by the policy, we analyze four outcome metrics including ball speed, spin rate after the shooting, clearance height over the net at the crossing point, and the estimated winning probability of each return via a trained random forest classifier based on landing ball states and height over the net using the professional player data \cite{liu2012new}. For each metric, we report the median value computed over all successfully returned shots as summarized in the following Fig.~\ref{fig:sim2real_sts_combined}.
The results indicate that, when evaluated on real-world data, the policy operating with the proposed real-world predictor consistently produces returns with higher ball speed and greater spin compared to those generated using a simulator. At the same time, the returned balls exhibit a lower clearance height over the net, producing flatter and more aggressive trajectories. Such shot characteristics are generally more difficult for human opponents to intercept and return.
These qualitative improvements are further reflected in the estimated winning probability (Win Prob.). In particular, for the Stroke-Uni and Stroke-Pro datasets under the 115–125 ms prediction horizon, the use of the proposed real-world predictor approximately doubles the winning probability relative to the simulator. Therefore, in real-world play, using the real-world predictor not only increases the success rate of returns but also leads to higher-quality and more competitive shots.

\begin{figure}[thpb]
    \centering
    \includegraphics[width=0.98\linewidth]{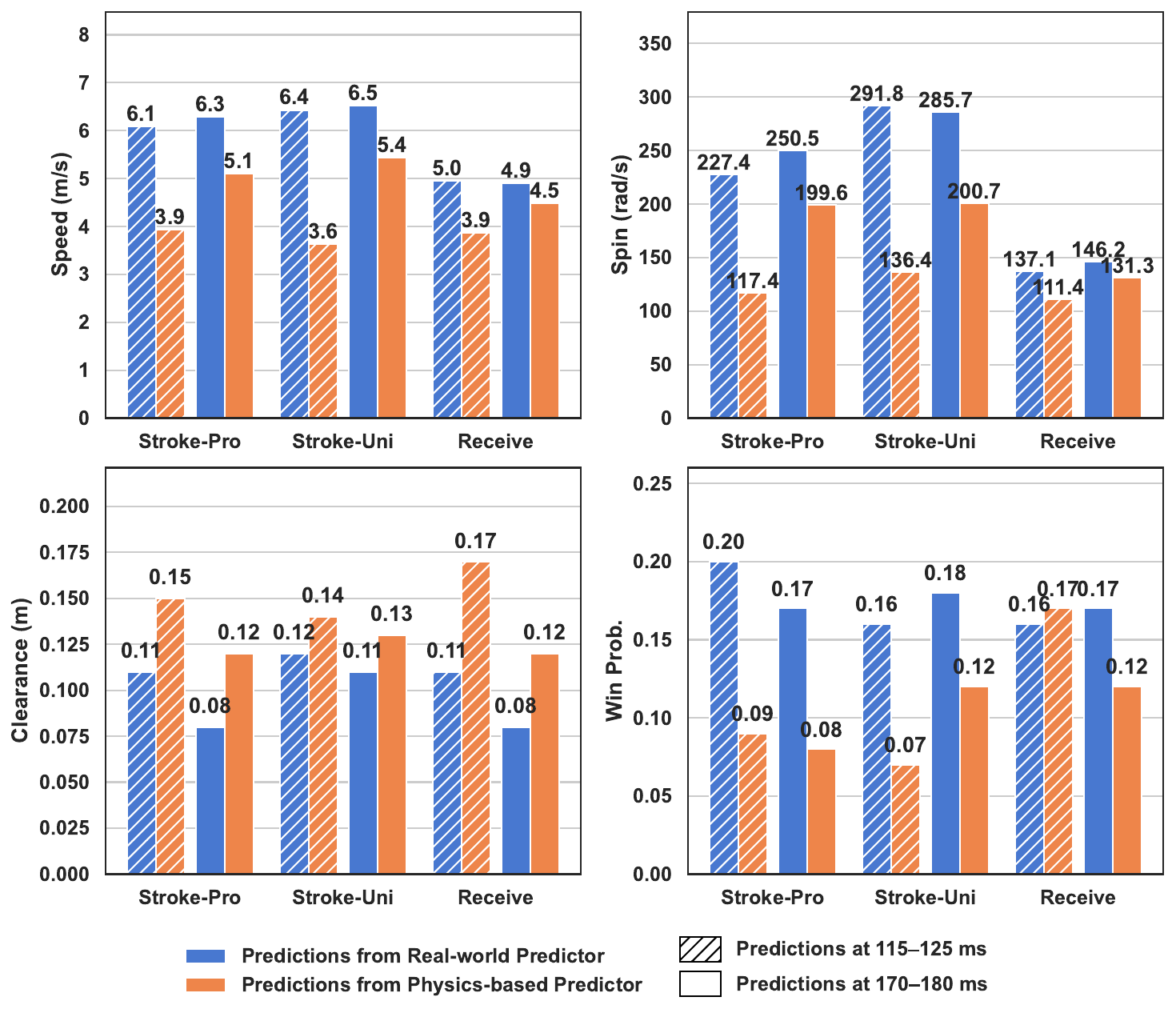} 
    \caption{Shot quality comparison: Median values of four metrics-ball speed, spin rate, net clearance, and estimated winning probability-are shown for successfully returned shots in three datasets (Stroke-Pro, Stroke-Uni, Receive). Policies using the real-world predictor (blue) are compared with those using the physics-based predictor (orange) at two prediction horizons (115–125 ms (hatched); 170–180 ms (solid)). Using the real-world predictor consistently produces faster, higher-spin, and lower-clearance returns, resulting in higher estimated winning probabilities.}
    \label{fig:sim2real_sts_combined}
\end{figure}

\section{CONCLUSIONS}
We presented a transformer-based approach for predicting the state of a table tennis ball, addressing limitations of physics-based methods that rely on accurate parameter identification and precise initial states, as well as prior learning-based approaches that struggle with long-range temporal dependencies. 
To support robust training and generalization, we collected a large-scale and diverse real-world dataset and released testing samples to facilitate standardized evaluation. 
We demonstrated the effectiveness of the proposed predictor through real-world deployment on a robotic arm, enabling stable and responsive human–robot rallies. 
Furthermore, we introduced a sim-to-real transfer strategy in which the predictor used during simulation training is replaced at deployment with the proposed predictor trained on real-world data. Experiments show that this approach preserves the efficiency and scalability of simulation-based training while mitigating the sim-to-real gap. 
Future work will explore extending this strategy to other robotic systems where learned predictors provide structured inputs to control policies.

\bibliographystyle{IEEEtran}
\bibliography{bib}

\end{document}